\documentclass[11pt]{article}

\usepackage[T1]{fontenc}
\usepackage[utf8]{inputenc} 
\usepackage[english]{babel}

\usepackage{lmodern}
\usepackage{microtype}
\usepackage{graphicx}
\usepackage{booktabs}
\usepackage{array}
\usepackage{tabularx}
\usepackage{ragged2e}
\usepackage{longtable} 
\usepackage{url}
\usepackage[hidelinks]{hyperref}
\usepackage{caption}
\usepackage{subcaption}
\usepackage{float}

\usepackage[margin=1in]{geometry}

\title{Teachy Mini: Development and Preliminary Evaluation of a Knowledge-Based Generative Social Robot for Higher Education}

\author{
Stephan Vonschallen$^{1,2}$ \and
Karim Kaufmann$^{1}$ \and
Dominique Oberle$^{1}$ \and
Friederike Eyssel$^{2*}$ \and
Theresa Schmiedel$^{1*}$
}

\date{} 

\begin{document}
\maketitle
\noindent
\begin{center}
$^1$Institute of Information Systems, Zurich University of Applied Sciences, Switzerland\\
$^2$Center for Cognitive Interaction Technology, Bielefeld University, Germany\\
$^*$Friederike Eyssel and Theresa Schmiedel share senior authorship\\[4pt]
\textbf{Corresponding author:} Stephan Vonschallen (\texttt{stephan.vonschallen@zhaw.ch})
\end{center}

\begin{abstract}

Generative social robots (GSRs) powered by large language models offer
new possibilities for personalized tutoring in higher education, but
also introduce risks related to misinformation, missing transparency, or
reinforcing incorrect student responses. Prior work identified
knowledge-based design (KBD) requirements that define the informational
prerequisites for GSRs to manifest responsible and effective tutoring
behavior in higher education. In this paper, we operationalized selected
KBD requirements in the Reachy Mini robot platform through system
prompting, retrieval-augmented generation, and stateful prompt
orchestration. As a result, we present Teachy Mini, a GSR tutoring
system that was developed using KBD. To test the system, we conducted a
preliminary evaluation study. Participants (N = 24) completed a
robot-guided learning session about research methodologies. They learned
either with Teachy Mini or with a control version that did not follow
KBD principles. Teachy Mini was perceived as significantly more aligned
with responsible tutoring behavior than the control robot. Moreover, a
manipulation check illustrated that Teachy Mini used personalization,
slide-grounded explanations, Socratic questioning, affective support,
and learner-anchored feedback more consistently than the control robot.
No significant between-condition differences were found in system
acceptance, intrinsic motivation, or learning effectiveness, although
exploratory analyses suggested a positive effect of KBD on objective
learning gains when accounting for learner preferences. Furthermore, an
exploratory path model suggested that the robot's perceived responsible
behavior may be associated with motivation, acceptance, and subjective
learning effectiveness. Overall, the study offered an initial
implementation and preliminary evaluation of KBD for GSR tutoring,
indicating that KBD can shape responsible robot behavior and potentially
increase learning effectiveness in robot-supported learning.
\end{abstract}

\section{Introduction}\label{introduction}

Teachers and students in higher education increasingly rely on
technology-supported learning to address the challenges of large class
sizes and increasing demands for self-directed study
\cite{ramdasCreatingEquitableLearning2025,vonschallenKnowledgeBasedDesignRequirements2026,youngTiredFailingStudents2021}.
While study groups and personal tutoring can support this process
through explanation, feedback, and social motivation
\cite{loesEffectCollaborativeLearning2022,tullisWhyDoesPeer2020,zhangEffectsPrivateTutoring2022},
their availability is often limited by time, access, and institutional
resources. Social robots offer a promising complementary approach, as
they can provide interactive learning support, foster engagement, and
create a sense of social presence
\cite{belpaemeSocialRobotsEducation2018,paiEffectivenessSocialRobots2024,wooUseSocialRobots2021}.
However, effective tutoring requires more than the delivery of factual
information: It also depends on nuanced social understanding, adaptive
communication, pedagogically appropriate feedback, and sensitivity to
learner motivation
\cite{tiszaGenerativeAIpoweredSocial2025,vonschallenKnowledgeBasedDesignRequirements2026}.

With the integration of large language models (LLMs) into social robots,
natural, human-like communication between social robots and human users
has become possible. LLM-driven \emph{generative social robots}
\cite{vonschallenKnowledgebasedDesignRequirements2026b} are capable of
open-ended, context-sensitive dialogue
\cite{billingLanguageModelsHumanrobot2023,kimUnderstandingLargeLanguageModel2024},
enabling them to respond flexibly to learner questions, provide
personalized explanations, and sustain flexible tutoring conversations
in real time \cite{cordova-esparzaAIpoweredEducationalAgents2025}. This
represents a substantial departure from earlier generations of tutoring
robots, whose limited dialogic flexibility constrained their educational
utility \cite{belpaemeSocialRobotsEducation2018}.

At the same time, the use of GSRs in education introduces substantial
risks: LLM-based systems generate responses probabilistically
\cite{benderDangersStochasticParrots2021}. Hence, they may produce
inaccurate or misleading explanations
\cite{chenWhenHelpfulnessBackfires2025,ciubotaruHallucinationProblemGenerative2025},
confirm incorrect statements made by learners because of sycophantic
behavior
\cite{chenWhenHelpfulnessBackfires2025,chengSycophanticAIDecreases2026,sunBeFriendlyNot2026},
decrease critical thinking
\cite{azeemPersonalityCorrelatesAcademic2025,zhaiEffectsOverrelianceAI2024},
or even discriminate against students
\cite{hundtLLMdrivenRobotsRisk2025}. Further concerns include
overreliance on AI-based tutoring
\cite{houMeasuringUndergraduateStudents2025,zhaiEffectsOverrelianceAI2024}
and privacy risks associated with the processing of student data
\cite{bakshOpensourceRoboticStudy2024,tiszaGenerativeAIpoweredSocial2025,williamsEthicalImplicationsUsing2024}.
These concerns are amplified in higher education, where students may
interact with such systems under evaluative pressure and may rely on
them when preparing for examinations or completing learning tasks.

The coexistence of these risks and opportunities highlights the need to investigate how the responsible deployment of GSRs in higher education can be facilitated.Within the scope of this work, responsible behavior refers to actions that align with the needs of students and teachers while also upholding broader social norms and ethical standards.To support responsible GSRs in higher education, we
previously identified Knowledge-Based Design (KBD) requirements for
tutoring-oriented GSRs used by university students for learning lecture
content \cite{vonschallenKnowledgeBasedDesignRequirements2026}. However,
these KBD requirements have not yet been implemented and empirically
tested in an educational context. Consequently, it remains unclear
whether these KBD requirements can be translated into a tutoring GSR and
whether such an implementation has a positive impact on learners. The
present research addresses this gap by developing and testing a GSR
tutoring prototype that operationalizes KBD requirements on a social
robot platform. Specifically, we developed Teachy Mini, a robot
tutoring system designed for single-session interactions with students. We then compared Teachy
Mini with a robot tutor that did not follow KBD guidelines, but had the
same hardware, conversational capabilities, and language model. Thereby,
we examined whether and how KBD can be integrated into a GSR tutor and
whether the integrated design features can shape students' perceptions of the robot
and learning experiences.

\section{Related Work}\label{related-work}

Social robots have been studied as tutors, peer learners, and learning
companions in educational settings
\cite{belpaemeSocialRobotsEducation2018,xuRoboticRolesEducation2025}.
Existing review articles suggest that their physical embodiment and
social cues in particular support engagement, motivation, and learning.
However, effects vary substantially across tasks, age groups, and study
designs
\cite{anwarSystematicReviewStudies2019,mubinREVIEWAPPLICABILITYROBOTS2013,wooUseSocialRobots2021}.
Most works study children in school contexts, while research in higher
education remains comparatively underrepresented
\cite{belpaemeSocialRobotsEducation2018}. In one of the few
university-level studies, Donnermann et al.
\cite{donnermannSocialRobotsApplied2022} evaluated an adaptive tutoring
robot for exam preparation and found that robot-supported tutoring was
feasible and positively perceived, but that adaptivity did not clearly
outperform a non-adaptive configuration. This suggests that social
robots may support self-directed learning in higher education, but that
their added value depends less on embodiment alone than on the quality
and configuration of their tutoring behavior.

Recent work in human-robot interaction has advanced beyond using
scripted or Wizard-of-Oz-controlled systems toward AI-based and
LLM-driven tutoring robots. LLMs can enable more flexible dialogue,
adaptive explanations, and open-ended interaction than prior rule-based
systems, making them attractive for tutoring scenarios that require
responsiveness to learner questions
\cite{kasneciChatGPTGoodOpportunities2023}. Hence, GSRs are increasingly
used in educational settings for personalized feedback and adaptive
content generation
\cite{ackermannHowAdaptiveSocial2025,bakshOpensourceRoboticStudy2024,lampropoulosSocialRobotsEducation2025}.
To illustrate, Smit et al. \cite{smitEnhancingEducationalDynamics2024}
evaluated a GSR in an educational setting and found that LLM integration
improved interaction quality and user satisfaction compared to scripted
alternatives. Similarly, Elgarf et al. \cite{elgarfFosteringChildrensCreativity2024} explored how LLM-driven
social robots may support collaborative learning interactions, showing
that generative AI enables robots to adapt dynamically to individual
learner input in ways that pre-programmed systems cannot.
However, empirical studies that investigate interactions with GSR tutors
and learners remain scarce, and evidence based on empirical research in
higher education is particularly limited.

Such lack of evidence is problematic because tutoring GSRs raise design
questions that go beyond usability or mere learning performance.
Responsible educational tutoring requires factual accuracy,
transparency, privacy protection, learner autonomy, pedagogical
appropriateness, and sensitivity to students' motivational and emotional
states
\cite{holmesEthicsAIEducation2022,kasneciChatGPTGoodOpportunities2023,vonschallenKnowledgeBasedDesignRequirements2026}.
These requirements are difficult to ensure in GSRs because their
responses are generated probabilistically and may include
hallucinations, unsupported claims, overconfident explanations, or
socially inappropriate guidance
\cite{kasneciChatGPTGoodOpportunities2023,kimUnderstandingLargeLanguageModel2024,weidingerEthicalSocialRisks2021}.
Existing design frameworks, such as \emph{AI for Social Good} or
\emph{Value Sensitive Design}
\cite{floridiHowDesignAI2020,umbrelloMappingValueSensitive2021}, provide
important principles for ethical AI use, but they do not specify what
specific knowledge a GSR must possess to express these ethical
principles through its behavior. This leaves a gap between abstract
responsible-AI principles and the concrete system configuration required
for responsible tutoring.

\begin{figure}[ht!]
\begin{center}
\includegraphics[width=\textwidth]{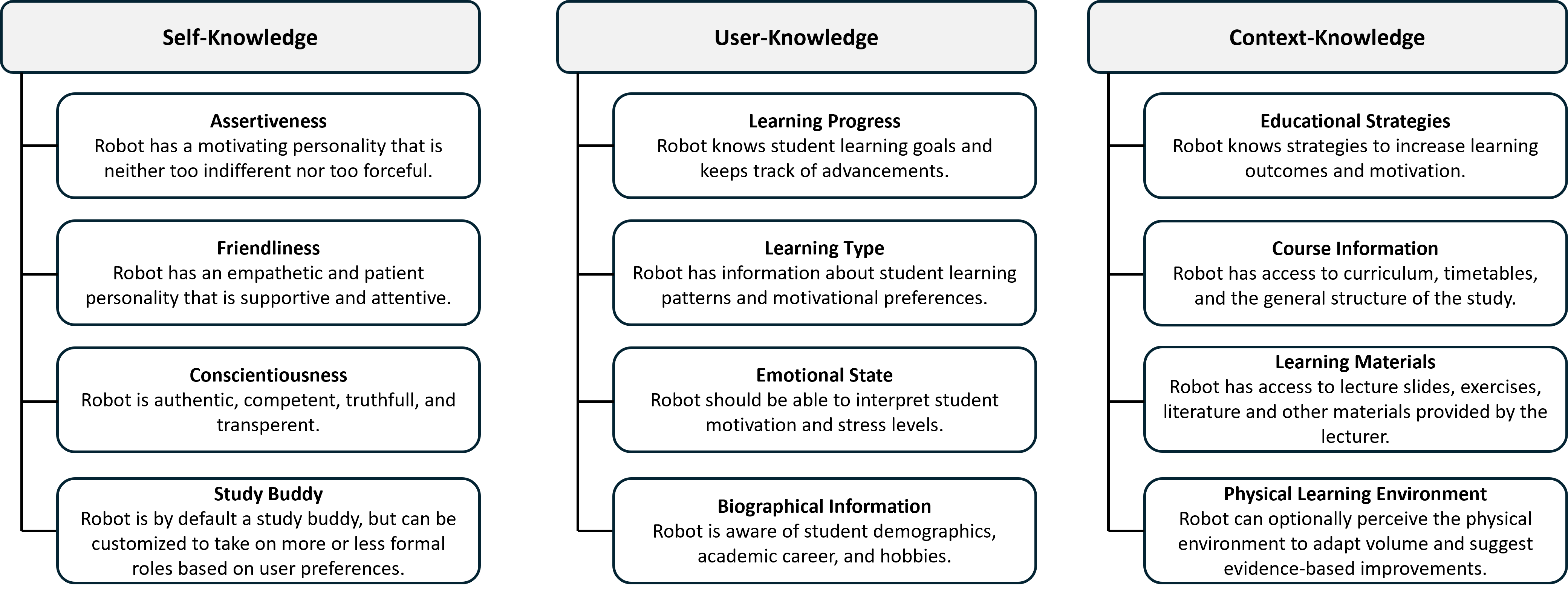}
\end{center}
\caption{Knowledge-based Design Requirements identified by
Vonschallen et al.
\cite{vonschallenKnowledgeBasedDesignRequirements2026}}
\label{Figure 1}
\end{figure}

To address this gap, Vonschallen et al.
\cite{vonschallenKnowledgeBasedDesignRequirements2026} conducted a
qualitative interview study with university students and lecturers to
identify desired behaviors and KBD requirements for effective and
responsible tutoring-oriented GSRs in higher education. The study was
based on the assumption that responsible robot behavior depends on the
availability of \emph{self-}, \emph{user-}, and
\emph{context-knowledge}. Regarding expected robot behavior, a tutoring
robot in higher education should have pedagogical skills to motivate and
teach students effectively. The robot should also give personalized
support that is tailored to learners' backgrounds and learning
preferences. Moreover, it was deemed important that the robot would give
correct information that is aligned with the content to be studied. The
results suggested that a robot should teach students in a transparent
way and mention the sources of the transmitted information. According to
these desired behaviors, KBD requirements further specified the type of
information necessary for the GSR to realize optimal responses
\cite{vonschallenKnowledgeBasedDesignRequirements2026}. Regarding
\emph{self-knowledge,} the robot's role should be adaptive, with the
default being a study buddy that is friendly, motivating, and
conscientious. To enable personalization, the robot's
\emph{user-knowledge} should encompass learner-specific information such
as learning goals, learning progress, motivation type, emotional state,
academic background, and preferred learning style, while respecting
privacy and informed consent. For context-sensitive educational support,
the robot should have \emph{context-knowledge} about official learning
materials, course-related information, educational strategies, and,
where appropriate, the physical learning environment. Figure \ref{Figure 1}
summarizes these KBD requirements.

\section{\texorpdfstring{Prototype Development and Knowledge Integration
}{Prototype Development and Knowledge Integration }}\label{prototype-development-and-knowledge-integration}

Building on the research of Vonschallen et al.
\cite{vonschallenKnowledgeBasedDesignRequirements2026}, the present work
integrates selected KBD requirements in a GSR tutor prototype, which we
called Teachy Mini. Our goal was to evaluate whether Teachy Mini would
lead to more responsible and effective tutoring interactions compared to
a robot that was not designed with KBD. Hence, Teachy Mini was not
intended as a fully deployable educational system, but as an
experimental implementation that makes it possible to examine how KBD
shapes tutoring behavior during a single learning interaction. The
prototype was implemented on the open-source Reachy Mini platform (Pollen Robotics\footnote{https://huggingface.co/spaces/pollen-robotics/Reachy\_Mini}),
a small desktop social robot that is suitable for individual learning
interactions. It supported spoken dialogue and provided simple embodied
social cues through head and antenna movements.

For Teachy Mini's communication system, we used an adaptation of the
\emph{Reachy Mini Conversation App}\footnote{\url{https://github.com/pollen-robotics/reachy_mini_conversation_app}}. This application supported real-time spoken
conversation using the OpenAI Realtime API (gpt-realtime-1), as well as
adaptive motion. These two features were important for enabling natural,
real-time human--robot interaction
\cite{studerusFrameworkLowLatencyLLMDriven2026}. We modified several of
the application's functions to increase the quality of the interaction.
This included recovery from incomplete model responses, setting higher
interruption thresholds, and removing potentially distracting non-verbal
idle movements during pauses. We then used three main methods to provide
the robot with the required knowledge: \emph{Dynamic system prompting}
\cite{ayyamperumalCurrentStateLLM}\emph{, retrieval-}augmented
generation \cite{kleselRetrievalAugmentedGenerationRAG2025}, and
\emph{stateful prompt orchestration}
\cite{wuPROMISEFrameworkModeldriven2024}. \emph{System prompting} gives
the system instructions about its overarching behavior and role.
\emph{Dynamic system prompting} uses a combination of stable
instructions and dynamic elements that can change within an interaction
to allow for personalization. \emph{RAG} is used to retrieve relevant
information from an external knowledge source and make it available to
the LLM when generating a response. This allows the model's outputs to
be grounded in task-specific material rather than relying only on its
general training data. Lastly, \emph{stateful prompt orchestration}
dynamically updates or supplements the model instructions during the
interaction based on different dialogue states, i.e., interactional
situations that require a more precise, pre-defined response from the
robot. For example, if a verbal pattern related to confusion, such as
\emph{``I don't understand this''}, is detected during a conversation, a
state-specific instruction is added to the user input: \emph{``Rephrase
your final thought in simple words using different vocabulary and a
concrete analogy. Do NOT repeat the same words.''} This was used to
increase the robustness of the robot's communication at critical moments
in a conversation.

\begin{figure}[ht!]
\begin{center}
\includegraphics[width=0.8\textwidth]{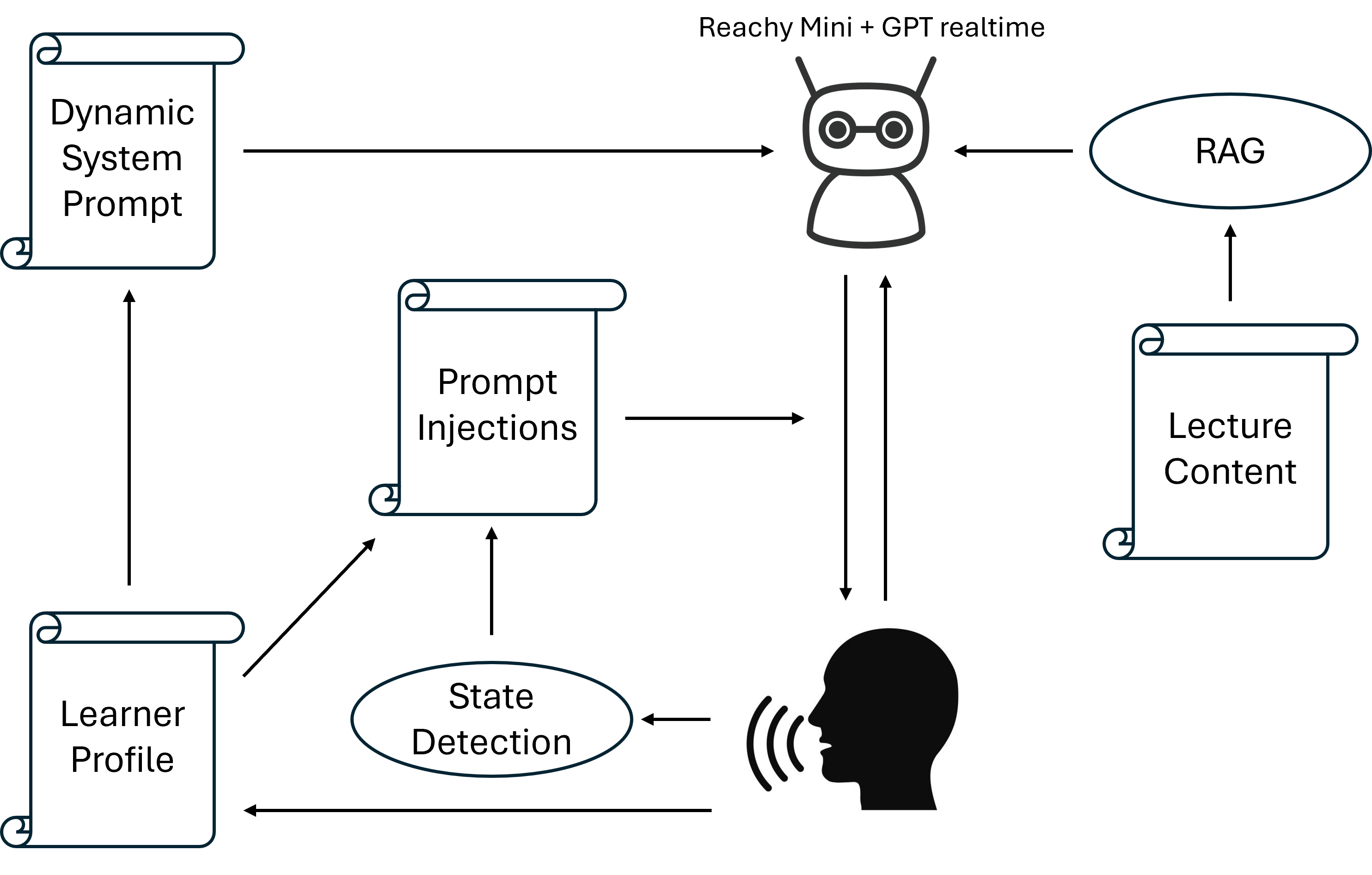}
\end{center}
\caption{Overview of Prompt Orchestration}
\label{Figure 2}
\end{figure}

Figure \ref{Figure 2} gives an overview of the system's architecture, with a focus on
prompt orchestration. More specifically, in the dynamic system prompt,
the model was given static instructions with dynamic variable
placeholders, the values of which were derived from a learner profile.
This learner profile was created at the beginning of the interaction
through an onboarding conversation with the robot. In parallel, RAG was
used to provide the model with lecture content. After receiving a
response from a human user, regular expressions -- that is, rule-based
text-matching patterns -- were used to detect the relevant
conversational state. In particular, the user's response was checked for
predefined verbal patterns indicating states such as frustration,
uncertainty, confusion, requests for content explanations, requests for
more depth, or requests that the robot stop asking questions. If such a
pattern was detected, the system injected a short state-specific
instruction into the next user prompt, for example instructing the robot
to acknowledge frustration, simplify an explanation, or provide a more
direct explanation. Depending on the state, these prompts were sometimes
dynamic, that is, they used placeholders whose values were drawn from
variables in the learner profile. For example, the robot could remind
students of their personal goals when they expressed frustration. A list
of all states and prompts is available on researchbox.org (ID: \#8129).
Based on this architecture, we integrated the KBD requirements into the
robot system. The implementation code is available on GitHub GitHub\footnote{\url{https://github.com/StephanVonschallen/Teachy-Mini}}.

Overall, we integrated a selection of the previously identified
\emph{self-knowledge}, \emph{user-knowledge}, and
\emph{context-knowledge} design requirements
\cite{vonschallenKnowledgeBasedDesignRequirements2026} into Teachy
Mini's interaction design. As Teachy Mini was designed for application
in an experimental single-session context, the design requirement
``\emph{course information''}, which is necessary for interactions over
longer periods of time, was not integrated. Similarly, the design
requirement ``\emph{learning progress''} was only partly accounted for
by integrating learner goals, but not actual progress over multiple
interactions. Moreover, although several role profiles were developed,
the preliminary evaluation used only the \emph{``study buddy''} profile,
which was widely preferred by students and lecturers
\cite{vonschallenKnowledgeBasedDesignRequirements2026}, to keep the
interaction role consistent across participants. Similarly, we did not
integrate the optional design requirement ``\emph{physical learning
environment''} as this may have introduced additional variability in
experimental settings, and because this design requirement only received
mixed support in the identification study due to privacy concerns about
being filmed and limited perceived usefulness
\cite{vonschallenKnowledgeBasedDesignRequirements2026}.

\emph{User-knowledge} was integrated to personalize Teachy Mini's
behavior to user preferences and needs. Most learner-related information
was acquired through a short onboarding interaction with the robot. In
this onboarding interaction, Teachy Mini was instructed to ask seven
opening questions to acquire relevant user information regarding the
design requirements ``\emph{biographical information''},
``\emph{learning type''}, and ``\emph{learning progress''} (Table \ref{tab1}).
The acquired user information from these opening questions was extracted
using rule-based regex parsing and stored in a user profile that was
added to the system prompt. The design requirement \emph{``emotional
state''} was integrated through state detection. Specifically, the robot
received information through prompt injection indicating whether the
user appeared frustrated, uncertain, or confused.

\begin{table}[H]
    \centering
    \caption{Opening Questions}
    \label{tab1}
    \renewcommand{\arraystretch}{1.25}
    \setlength{\tabcolsep}{6pt}

    \begin{tabularx}{\textwidth}{
        >{\raggedright\arraybackslash}p{0.08\textwidth}
        >{\raggedright\arraybackslash}X
        >{\raggedright\arraybackslash}p{0.24\textwidth}
    }
        \toprule
        \textbf{Question} & \textbf{Content} & \textbf{Design Requirement} \\
        \midrule

        Q1 &
        What is your name? How would you like me to address you? &
        Biographical Information \\

        Q2 &
        What are you studying, and which semester are you in? &
        Biographical Information \\

        Q3 &
        What motivates you when learning? What drives you? &
        Learning Type \\

        Q4 &
        How do you prefer to learn: in a structured way with clear steps,
        or through a more exploratory approach? &
        Learning Type \\

        Q5 &
        What do you enjoy doing in your free time? What are your hobbies or
        interests? &
        Biographical Information \\

        Q6 &
        What kind of learning atmosphere do you prefer: strictly factual,
        or one that also includes some humor? &
        Role \\

        Q7 &
        What would you like to achieve in today's session?
        Do you have a specific learning goal? &
        Learning Progress \\

        \bottomrule
    \end{tabularx}
\end{table}

\emph{Context-knowledge} was integrated to ensure that Teachy Mini's
behavior was sensitive to the educational context. To fulfill the KBD
requirement \emph{``learning materials''}, we implemented an upload
mechanism for PDF documents, where users could upload lecture slides by
drag and drop. The system could access these PDFs using RAG, providing
it with context-sensitive topic knowledge. This implementation is
similar to functions offered by chat interfaces of popular generative AI
systems such as ChatGPT and Gemini, but these functions are less
commonly used for HRI implementations such as the baseline \emph{Reachy
Mini Conversation App}. To apply \emph{``educational strategies'',} we
used dynamic system prompting and stateful prompt orchestration. The
robot was instructed to use scaffolding strategies, including Socratic
questioning (i.e., questions that guide learners to examine their
reasoning and think for themselves), targeted hints, simplified
explanations, and corrective feedback. Concretely, the system prompt
included instructions to ask an in-depth question after each concept was
explained to check the students' level of understanding. These questions
were content-sensitive, e.g., \emph{``Which point differentiates
concepts A and concept B the most?",} or \emph{``Which two properties
did we just discuss?{\kern0pt}``.} Further, when a student's answer was
correct or partially on track, the robot should give one specific
sentence of acknowledgement that names what the student understood,
avoid generic confirmations, and vary its wording across turns. When a
student gave an incorrect answer, the robot was tasked to avoid direct
rejection, briefly acknowledge the reasoning attempt, identify what is
partially correct or where the reasoning diverges, ask one targeted
guiding question, and only provide a small hint after a second failed
attempt. However, if the user wanted to engage in a different learning
approach that did not include follow-up questions, the robot was
instructed via system prompt to follow the user's suggested approach.
Additionally, stateful prompt orchestration was applied to ensure an
adaptive use of the educational strategies. To illustrate, when the
student explicitly asked for more depth, prompt injection was used to
temporarily stop the robot from Socratic questioning. Instead, the robot
was instructed to provide a detailed explanation with concrete
terminology, add a specific example, and then ask one
comprehension-check question before returning to interactive
questioning. When the student explicitly asked the robot to stop asking
questions or to simply explain or summarize the content, the robot's
system prompt was updated to suspend Socratic questioning, provide
direct explanations without follow-up questions for the next few turns,
and then return to its default interactive tutoring style unless the
student repeated the request. Lastly, if the user was detected to be
confused about an explanation, the robot was instructed to restate its
previous point in simpler language, use different wording and a concrete
analogy, and avoid repeating the same explanation.

\emph{Self-knowledge} was integrated by defining Teachy Mini's role,
interaction style, didactic stance, and behavioral boundaries. The
design requirement \emph{``friendliness''} was implemented directly
through \emph{system prompting}, instructing the robot to communicate in
a warm, friendly, encouraging, and supportive tone. Furthermore, if the
state detection mechanism identified signs of user frustration, Teachy
Mini was prompted to react in an appreciative manner, reminding the user
of their stated goal and motivation derived from the user profile (Table
1, Q7). ``\emph{Conscientiousness''} was integrated through system
prompts that required the robot to communicate carefully, transparently,
and truthfully. Teachy Mini was instructed not to invent facts, sources,
studies, or citations, to explicitly acknowledge uncertainty when
information was unclear, and to distinguish between slide-grounded
information and general model knowledge. When referring to course
material, the robot was required to use the retrieval tool and cite the
relevant slide rather than claiming unsupported slide content.
\emph{``Conscientiousness''} was also reflected in identity and safety
rules: The robot had to identify itself as an AI-powered Reachy Mini
rather than pretending to be human, state when no camera was available,
and refer students to appropriate counseling services in cases of severe
personal distress. The design requirement \emph{``assertiveness''} was
operationalized in combination with \emph{``educational strategies''}
\emph{(context-knowledge)}. The robot was instructed to guide the
interaction through scaffolding and concept-specific follow-up questions
rather than waiting passively for user requests. At the same time, this
initiative was constrained by pacing and autonomy rules: When user
states were detected that indicated time pressure, beginner status,
frustration, or a preferred learning approach, the robot was instructed
to reduce questioning, provide more direct support, or follow the
student's chosen procedure. Lastly, the design requirement \emph{``study
buddy''} was integrated through system prompting. Techy Mini was
assigned a peer-level tutoring role and explicitly distinguished from an
authoritative teacher or examiner. This role positioned the robot as a
learning companion that supports students in working through the
material while leaving responsibility and control with the learner. To
account for the broader requirement of role adaptability, we also
created alternative role profiles in which the robot could assume more
assertive roles, such as a coach or professor.

\section{Preliminary Evaluation}\label{preliminary-evaluation}

Teachy Mini was evaluated in an interaction study to examine whether
integrating KBD requirements changed how students experienced and
interacted with a GSR tutor. Accordingly, the study compared a KBD
version of the robot with a control version that did not follow KBD
guidelines, but operated on the same platform and with the same LLM. The
study was approved by the ethics committee of Bielefeld University (ID:
EUB-2025-369) and preregistered on aspredicted.org (ID: 287,000).

Four hypotheses were derived from prior work on educational social
robots, AI-based tutoring, and KBD. First, educational technology
acceptance research suggests that learners are more likely to accept a
system when it is perceived as useful, easy to interact with, and
appropriate for the learning task
\cite{davisPerceivedUsefulnessPerceived1989,heerinkAssessingAcceptanceAssistive2010}.
Because KBD provides the robot with a consistent tutoring role, access
to learner-related information, and course-specific content, we expected
Teachy Mini to be perceived as more acceptable than the control robot:

\emph{H1: Students interacting with Teachy Mini will report greater
system acceptance than students interacting with the control robot.}

Second, prior research on social robots and AI-based learning systems
suggests that personalization, adaptive feedback, and socially engaging
interaction can support learner motivation
\cite{belpaemeSocialRobotsEducation2018,donnermannSocialRobotsApplied2022,kasneciChatGPTGoodOpportunities2023}.
Since Teachy Mini was designed to use learner-specific information,
provide motivating feedback, and adapt its tutoring behavior to the
student's needs, we expected it to increase students' intrinsic learning
motivation.

\emph{H2: Students interacting with Teachy Mini will report greater
intrinsic learning motivation than students interacting with the control
robot.}

Third, intelligent tutoring and robot-assisted learning research
suggests that learning can benefit when support is aligned with the
learner's current understanding and grounded in the relevant
instructional material
\cite{baillifardEffectiveLearningPersonal2025,liRetrievalaugmentedGenerationEducational2025,tasdelenGenerativeAIClassroom2025}.
Teachy Mini was designed to support learning through slide-grounded
explanations, scaffolding, and adaptive questioning. We therefore
expected it to support learning more effectively than the control robot.

\emph{H3: Students interacting with Teachy Mini will demonstrate
greater objective (H3a) and subjective (H3b) learning effectiveness than
students interacting with the control robot.}

Finally, the central assumption of the KBD framework is that responsible
tutoring behavior depends on the knowledge available to the robot. A
robot that knows its role and boundaries, can use relevant learner
information, and is grounded in the educational context should be better
able to behave in ways that are transparent, pedagogically appropriate,
and responsive to student needs
\cite{vonschallenKnowledgeBasedDesignRequirements2026}. We therefore
expected students to perceive Teachy Mini as more aligned with
responsible robot tutoring behavior.

\emph{H4: Students interacting with Teachy Mini will perceive its
behavior as more closely aligned with responsible robot tutoring than
students interacting with the control robot.}

\subsection{Sample}\label{sample}

Participants were recruited through convenience sampling. This sampling
strategy was considered appropriate because the study was exploratory
and aimed to provide preliminary evidence. Participants were required to be currently enrolled at a university or to have
completed a university degree within the previous four years.
Participants also had to be familiar with lecture-based learning formats
and sufficiently proficient in German to complete the spoken interaction
with the robot, as all instructions, questionnaires, and robot interactions were conducted in German. Individuals
who were in a current student--teacher relationship with a member of the
research team were excluded.

The final sample consisted of 24 participants. Data were collected
between March and April 2026. Participants had a mean age of 31 years
(\emph{SD} = 4.31), ranging from 23 to 36 years. Sixteen participants
identified as male (66.7\%) and eight as female (33.3\%). Half of the
participants were currently enrolled students, whereas the other half
were not currently enrolled but had completed their degree within the
previous four years. The most common field of study was information
systems (\emph{n} = 14), with other subjects being law (\emph{n} = 2),
engineering (\emph{n} = 3), social sciences (\emph{n} = 2), design
(\emph{n} = 2), and architecture (\emph{n} = 1). Accordingly, the level
of education in the sample was relatively high, with four participants
who had completed secondary education, 13 participants with a bachelor's
degree, and seven participants with a master's degree. Only one of the
24 participants had previously interacted with a social robot.

\subsection{Design}\label{design}

The study applied a two-group between-subjects design. Accordingly,
participants interacted with one of two versions of the same Reachy Mini
tutoring prototype: A KBD condition and a control condition. In the KBD
condition, the Teachy Mini robot was used. This robot was configured
with the implemented KBD requirements, including structured
\emph{self-knowledge}, \emph{user-knowledge}, and
\emph{context-knowledge}. It used the study-buddy role profile,
personalized the interaction based on onboarding information, applied
scaffolding and Socratic questioning, and had access to the lecture
slides through retrieval-augmented generation.

In the control condition, participants interacted with the same Reachy
Mini robot platform that had identical communication capabilities and
used the same OpenAI Realtime language model. However, the control robot
did not have the structured KBD configuration. The robot used a neutral
and factual interaction style, did not personalize tutoring based on
the opening questions, did not use the study-buddy role framing or
KBD-specific scaffolding rules, and did not access the lecture slides
through the retrieval mechanism. Participants were randomly assigned to
either the KBD (\emph{n} = 12) or control (\emph{n} = 12) conditions
using equal-allocation randomization.

\subsection{Materials}\label{materials}

The materials for the current study included learning materials, a
knowledge test, and several scale measures. All study materials are
available on researchbox.org (ID: \#8129). The learning materials
consisted of 11 lecture slides on research methodologies for
\emph{Information Systems Research}. This topic was selected because it
falls within the teaching expertise of one member of the research team,
which allowed this researcher to assess whether the agent provided
pedagogically and factually accurate information. The slides introduced
core concepts of the topic, including definitions, distinctions between
research types, theory types, behavioral science and design science,
quantitative and qualitative methods, and research design decisions. The
same slide set was used in both conditions. During the tutoring session,
participants received the slides as printed A4 handouts. In the KBD
condition (Teachy Mini), the slide content was available to the robot
through retrieval-augmented generation, allowing the robot to ground its
explanations in the provided learning materials.

To assess students objective learning effectiveness (Hypothesis H3a), we
developed a 12-item multiple-choice knowledge test on \emph{Information
Systems Research} (Table \ref{tab2}). The knowledge test was designed for
pre-post test comparison. The items were constructed following the
revised Bloom taxonomy \cite{andersonTaxonomyLearningTeaching2009}, with
a balanced distribution of items across the cognitive process dimensions
\emph{recall}, \emph{comprehension}, and \emph{transfer}. \emph{Recall}
refers to retrieving specific information that was learned,
\emph{comprehension} to understanding and explaining concepts, and
\emph{transfer} to applying knowledge to new situations. Each item was
presented as a single-choice question with four answer options.

\begin{table}[H]
    \centering
    \caption{Self-developed Knowledge Test}
    \label{tab2}
    \scriptsize
    \renewcommand{\arraystretch}{1.08}
    \setlength{\tabcolsep}{4pt}

    \begin{tabularx}{\textwidth}{
        >{\raggedright\arraybackslash}p{0.07\textwidth}
        >{\raggedright\arraybackslash}X
        >{\centering\arraybackslash}p{0.16\textwidth}
        >{\centering\arraybackslash}p{0.09\textwidth}
    }
        \toprule
        \textbf{Item} &
        \textbf{Description} &
        \textbf{Type} &
        \textbf{\(p\)} \\
        \midrule

        Q1\textsuperscript{*} &
        What is Information Systems Research primarily concerned with? &
        Recall &
        87.5\% \\

        Q2\textsuperscript{*} &
        Which statement correctly describes basic research? &
        Recall &
        95.8\% \\

        Q3 &
        What is the primary goal of Behavioral Science? &
        Recall &
        45.8\% \\

        Q4\textsuperscript{*} &
        Which characteristics define qualitative research? &
        Recall &
        83.3\% \\

        Q5 &
        A researcher develops a robot prototype and evaluates its usefulness
        for care facilities. Which paradigm does this work belong to? &
        Transfer &
        58.3\% \\

        Q6 &
        A team first conducts focus groups and then uses the findings to
        design a quantitative survey. What is this approach called? &
        Comprehension &
        75.0\% \\

        Q7 &
        Why is Information Systems Research described as
        interdisciplinary? &
        Comprehension &
        79.2\% \\

        Q8\textsuperscript{*} &
        What should the research design primarily be based on? &
        Comprehension &
        91.7\% \\

        Q9 &
        A study uses a questionnaire to compare whether a new robot prototype
        is better accepted than an existing system. In which quadrant of the
        methodological spectrum does this study fall? &
        Transfer &
        12.5\% \\

        Q10\textsuperscript{*} &
        One research team investigates fundamental patterns in communication
        between people and IT systems without developing specific
        recommendations for improvement. A second team uses these findings to
        improve a chatbot for care homes. How should these two projects be
        classified? &
        Transfer &
        83.3\% \\

        Q11 &
        A study investigates how students use a learning app in everyday life
        and what problems arise. Which characteristic of Information Systems
        Research is illustrated here? &
        Transfer &
        54.2\% \\

        Q12 &
        A researcher wants to understand why employees reject a new system.
        Which method is most suitable? &
        Comprehension &
        75.0\% \\

        \bottomrule
    \end{tabularx}

    \vspace{0.4em}

    \begin{minipage}{\textwidth}
        \scriptsize
        \textit{Note.} Items were translated from German.
        \(p\) = item difficulty index, calculated as the number of correct
        responses divided by the total number of responses. Higher values
        indicate lower difficulty. \textsuperscript{*}Items Q1, Q2, Q4, Q8,
        and Q10 were later removed; see Section \ref{analysis}.
    \end{minipage}
\end{table}

Several scale measures were used to assess all other outcome variables.
System acceptance as measured using the \emph{Technology Acceptance
Model} scale \cite{davisPerceivedUsefulnessPerceived1989}, covering
perceived usefulness, perceived ease of use, and intention to use.
Intrinsic learner motivation was measured with the
\emph{Interest/Enjoyment} subscale of the \emph{Intrinsic Motivation
Inventory} \cite{ryanIntrinsicExtrinsicMotivations2000}\emph{.}
Subjective learning effectiveness was assessed through the
\emph{Perceived Competence} subscale of the \emph{Intrinsic Motivation
Inventory} \cite{ryanIntrinsicExtrinsicMotivations2000}. To assess
whether participants perceived the robot's behavior as aligned with the
previously identified KBD requirements, we further developed a
\emph{Responsible Behavior} scale. The scale was derived from the
identified KBD requirements by Vonschallen et al.
\cite{vonschallenKnowledgeBasedDesignRequirements2026} and captured
whether the robot was perceived as context-sensitive, motivating,
personalized, transparent, friendly, non-intrusive, competent, and
trustworthy (Table \ref{tab3}). Items were answered on a 7-point Likert scale
ranging from 1 \emph{(``I do not agree at all'')} to 7 \emph{(``I fully
agree'').} Negatively worded items were reverse-coded where applicable.

\begin{table}[H]
    \centering
    \caption{Self-developed Perceived Responsible Behavior Scale}
    \label{tab3}
    \renewcommand{\arraystretch}{1.2}
    \setlength{\tabcolsep}{6pt}

    \begin{tabularx}{\textwidth}{
        >{\raggedright\arraybackslash}p{0.11\textwidth}
        >{\raggedright\arraybackslash}X
        >{\centering\arraybackslash}p{0.07\textwidth}
        >{\centering\arraybackslash}p{0.07\textwidth}
    }
        \toprule
        \textbf{Item} &
        \textbf{Description} &
        \textbf{\textit{r}} &
        \textbf{\(\lambda\)} \\
        \midrule

        E3\_1 &
        \textit{The robot was competent.} &
        .647 &
        .728 \\

        E3\_2 &
        \textit{The robot was trustworthy.} &
        .666 &
        .638 \\

        E3\_3 &
        \textit{The robot was friendly.} &
        .680 &
        .799 \\

        E3\_4\textsuperscript{*†} &
        \textit{The robot was intrusive.} &
        .232 &
        .196 \\

        E3\_5 &
        \textit{The robot tried to motivate me to learn.} &
        .601 &
        .651 \\

        E3\_6 &
        \textit{The robot was able to offer me support that was tailored to me.} &
        .799 &
        .836 \\

        E3\_7 &
        \textit{The robot took my personal background into account.} &
        .741 &
        .798 \\

        E3\_8\textsuperscript{*} &
        \textit{The robot spread false information.} &
        .549 &
        .566 \\

        E3\_9 &
        \textit{The robot stated where the information it conveyed came from.} &
        .666 &
        .626 \\

        E3\_10 &
        \textit{The robot's information was specifically tailored to the distributed lecture slides.} &
        .608 &
        .619 \\

        \bottomrule
    \end{tabularx}

    \vspace{0.5em}

    \begin{minipage}{\textwidth}
        \footnotesize
        \textit{Note.} Items were translated from German;
        \(r\) = corrected item--total correlation;
        \(\lambda\) = factor loading;
        * Items E3\_4 and E3\_8 were reverse-coded.† Item E3\_4 was later removed; see Section \ref{analysis}
    \end{minipage}
\end{table}

Apart from the dependent variables, several potential confounders were
measured, including gender (\emph{``male''}, \emph{``female''},
\emph{``diverse''}), age (in years), level of education
(\emph{``primary''}, \emph{``secondary''}, \emph{``bachelor''},
\emph{``master''}, \emph{``promotion''})\emph{,} previous interactions
with social robots (\emph{``yes''} / \emph{``no''}), preference for
learning in groups versus alone (with a self-developed 6-item scale with
items such as \emph{``I prefer learning alone''}, \emph{``I prefer
learning with others'', ``I'm more efficient when learning alone''}),
perceived baseline competence
\cite{ryanIntrinsicExtrinsicMotivations2000}, general learning
motivation \cite{moeltnerValidierungDeutschenUebersetzung2024}, and
general attitudes towards social robots
\cite{spatolaAttitudesRobotsMeasure2023}.

\subsection{Procedure}\label{procedure}

\begin{figure}[ht!]
\begin{center}
\includegraphics[width=\textwidth]{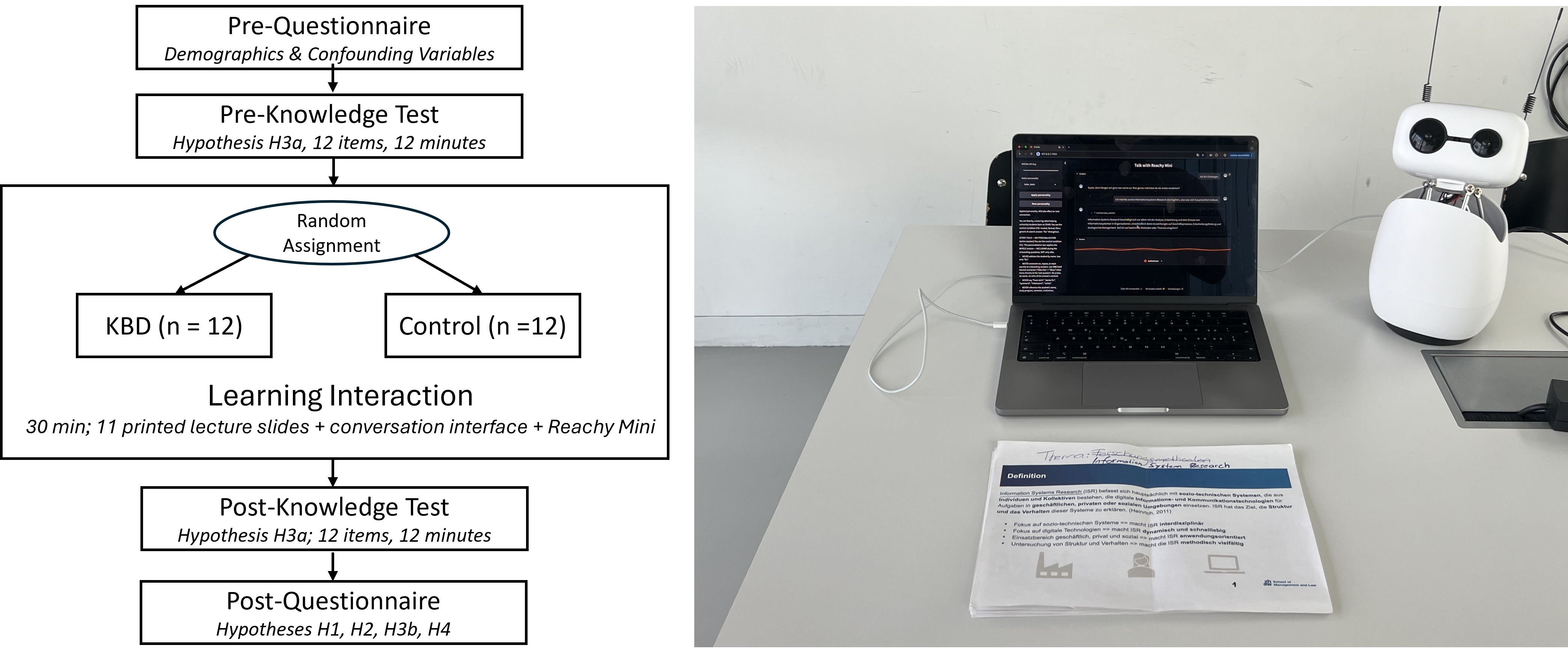}
\end{center}
\caption{Study Procedure and Setup}
\label{Figure 3}
\end{figure}

Figure \ref{Figure 3} gives an overview of the study procedure and setup. Each session was
conducted individually with one participant. After arrival, participants
received a brief introduction by a study supervisor and were informed
that the aim was to evaluate a robot tutoring system for university
learning. After providing written informed consent, participants
completed a pre-questionnaire assessing demographic information, prior
experience with social robots, attitudes toward robots, baseline
motivation, perceived competence regarding research methods, and
learning preferences. They then completed the 12-item knowledge pretest
on \emph{Information Systems Research} with a time limit of 12 minutes.

Participants were then randomly assigned to one of the two experimental
conditions and took part in a 30-minute tutoring interaction with the
Reachy Mini platform. During the session, the robot was positioned next
to a laptop displaying a live transcript of the interaction (Figure \ref{Figure 3}).
Participants also received the printed lecture slides, which were placed
face down before the tutoring session and turned over only when the
interaction began. The participants were instructed to use the session
to learn and understand the slide contents with the robot's assistance.
All interactions took place through spoken dialogue in German. During
the interaction and when participants completed surveys, the study
supervisor was not present in the room to avoid observer bias
\cite{mccambridgeSystematicReviewHawthorne2014}.

The full tutoring interaction was logged as a transcript through the
experimenter interface; audio was additionally recorded as a backup.
Before participants completed further tests and questionnaires, the
robot was removed from the room to reduce the possibility that the
robot's presence would influence participants' responses, consistent
with research on social responses to artificial agents
\cite{nassComputersAreSocial1994}. Participants then completed an
identical 12-item knowledge posttest on \emph{Information Systems
Research}, again with a time limit of 12 minutes. Afterward,
participants completed a post-interaction questionnaire assessing system
acceptance, intrinsic learning motivation, perceived competence,
perceived alignment with responsible robot behaviors, and open-ended
feedback. The open-ended questions asked what participants liked about
the interaction, what could be improved, how learning with the robot
differed from learning with a human tutor, and whether any technical
problems occurred. The study had an average duration of 74 minutes
(\emph{SD} = 20 minutes).

\subsection{Manipulation Check}\label{manipulation-check}

To check whether the robot's behavior meaningfully differed across
experimental conditions, we conducted a qualitative analysis of the
interaction transcripts after data collection. We randomly selected
interaction transcripts from three KBD interactions (KBD.1--3) and three
control interactions (CTRL.1--3) for qualitative content analysis
\cite{mayringQualitativeContentAnalysis2021}. The method was chosen for
its capacity to identify behavioral patterns systematically along
theoretically derived categories. As only a subset of the interaction
transcripts were coded, this analysis serves primarily an illustrative
purpose to demonstrate how the conditions differed.

The coding was conducted by one member of the research team and reviewed
by a second member of the research team, who is a lecturer in
Information Systems research methodologies. We applied a consensus-based
coding approach in which we discussed disagreements about the codes
until consensus was reached
\cite{braunSuccessfulQualitativeResearch2013}. The overarching main
categories of \emph{self-,} \emph{user-,} and \emph{context-knowledge}
were deductively added based on the previous conceptualization of these
knowledge types
\cite{vonschallenUnderstandingPersuasiveInteractions2026a}. The
subcategories \emph{``biographical information'', ``learning
preference'', ``learning goal'', ``learner emotion'', ``assertiveness'',
``friendliness'', ``conscientiousness'',} \emph{``role as study
buddy'',} \emph{``learning materials''}, and \emph{``educational
strategies''} were deductively added based on the selected design
requirements from the identification study
\cite{vonschallenKnowledgeBasedDesignRequirements2026}. Codes that
indicated whether these design requirements were present or absent were
inductively added from the material.

The main categories captured \emph{user-knowledge}, such as personal
address, references to hobbies or study background\emph{;
context-knowledge}, such as explicit slide references and slide-grounded
explanations, as well as Socratic questions as educational strategies;
and \emph{self-knowledge}, such as conscientiously providing correct
information, motivational statements, and affirmations. Affirmation was
further differentiated into generic praise and personalized feedback.
Similarly, for the design requirement \emph{``learning materials''}, we
compared sourced claims that refer to the lecture slides with claims
that did not have a source. Three of the ten deductively added
subcategories were removed after the coding \emph{(``learner goal'',
``learner emotion'', ``role as study buddy''}), because we were not able
to identify text passages that specifically related to them. In total,
655 text segments were coded (KBD: 469, Control: 186) into three main
categories and seven subcategories (Table \ref{tab4}). The coded documents had
similar interaction lengths: The KBD conditions had 164 robot messages
in total, while the control conditions had 163 messages in total.

\begin{table}[p]
    \centering
    \caption{Overview of Codes and Their Related Design Requirements}
    \label{tab4}
    \scriptsize
    \renewcommand{\arraystretch}{1.12}
    \setlength{\tabcolsep}{3pt}

    \begin{tabularx}{\textwidth}{
        >{\raggedright\arraybackslash}p{0.23\textwidth}
        >{\raggedright\arraybackslash}p{0.27\textwidth}
        >{\raggedright\arraybackslash}X
    }
        \toprule
        \textbf{Category and Code} &
        \textbf{Description} &
        \textbf{Example} \\
        \midrule

        \multicolumn{3}{l}{\underline{Biographical Information}} \\

        Name\newline \textit{[KBD: 86; CTRL: 0]} &
        Robot addresses the learner by name. &
        \textit{``Exactly, [name redacted], you've hit the core!''} \\

        Background\newline \textit{[KBD: 12; CTRL: 0]} &
        Robot refers to the learner's field of study or profession. &
        \textit{``In your degree program in Business Information Systems,
        this is typically done in projects such as the implementation of
        new IT systems.''} \\

        Hobby\newline \textit{[KBD: 5; CTRL: 0]} &
        Robot refers to the learner's hobby. &
        \textit{``Imagine it like strength training: not just theory about
        muscle growth, but actually doing it.''} \\

        \multicolumn{3}{l}{\underline{Learning Preferences}} \\

        Learning Procedure\newline \textit{[KBD: 5; CTRL: 0]} &
        Robot adjusts its tutoring style based on learner preferences. &
        \textit{``How would you like to proceed---go through the slides one
        by one, start with an overview, or jump straight into practice
        questions?''} \\

        Humor Preference\newline \textit{[KBD: 1; CTRL: 0]} &
        Robot uses humor in its explanations. &
        \textit{``It sounds more complicated than it actually is---kind of
        like the offside rule [from soccer].''} \\

        \multicolumn{3}{l}{\underline{Friendliness}} \\

        Affirmation with Explanation\newline
        \textit{[KBD: 90; CTRL: 13]} &
        Robot gives an affirmative response and explains why the answer was
        correct. &
        \textit{``Correct, you've recognized that the two methodologies are
        connected, but differ in their fundamental approach.''} \\

        Affirmation without Explanation\newline
        \textit{[KBD: 1; CTRL: 3]} &
        Robot gives an affirmative response without providing a reason. &
        \textit{``Great, then let's move straight on to the next concept!''} \\

        Patience\newline \textit{[KBD: 6; CTRL: 4]} &
        Robot expresses calmness and composure. &
        \textit{``All right, take your time with your notes. I'll be ready
        whenever you are.''} \\

        Joy\newline \textit{[KBD: 6; CTRL: 2]} &
        Robot expresses joy regarding student progress. &
        \textit{``I'm glad I could help you!''} \\

        \multicolumn{3}{l}{\underline{Assertiveness}} \\

        Correction\newline \textit{[KBD: 10; CTRL: 1]} &
        Robot corrects or clarifies a wrong or partially wrong statement. &
        \textit{``Your idea about basic research is a good one. However,
        what we're aiming for here is slightly different: The slide places
        particular emphasis on qualitative and quantitative approaches.''} \\

        No Correction\newline \textit{[KBD: 5; CTRL: 1]} &
        Robot does not correct a wrong or partially wrong statement. &
        \textit{``Good point---you've already identified that applied
        research moves toward concrete measurements.''}
        [After the participant suggested that applied research is about
        measuring outcomes.] \\

        Motivation\newline \textit{[KBD: 4; CTRL: 0]} &
        Robot provides encouragement. &
        \textit{``All right, if you have any more questions, I'm here to
        help. You're doing a great job!''} \\

        \multicolumn{3}{l}{\underline{Conscientiousness}} \\

        Ask for Confirmation\newline \textit{[KBD: 41; CTRL: 21]} &
        Robot checks whether the learner understood the explanation. &
        \textit{``Is this point clear now?''} \\

        False Information\newline \textit{[KBD: 0; CTRL: 10]} &
        Robot provides incorrect information. &
        \textit{``Yes, exactly. For example, analysis can investigate how
        user behavior develops and which factors influence it.''}
        [Mistaken for an explanation.] \\

        \multicolumn{3}{l}{\underline{Learning Materials}} \\

        Sourced Claim\newline \textit{[KBD: 66; CTRL: 0]} &
        Robot explicitly refers to a slide. &
        \textit{``The slide states that applied research deals with a
        phenomenon with the aim of improving something or putting it into
        practice.''} \\

        Claim without Source\newline \textit{[KBD: 15; CTRL: 130]} &
        Robot does not state the source of its explanation. &
        \textit{``Interviews can also be analyzed quantitatively.''} \\

        \multicolumn{3}{l}{\underline{Educational Strategies}} \\

        Socratic Question\newline \textit{[KBD: 116; CTRL: 1]} &
        Robot uses Socratic questions. &
        \textit{``How would you tell whether a piece of research is more
        qualitative than quantitative?''} \\

        \bottomrule
    \end{tabularx}

    \vspace{0.4em}

    \begin{minipage}{\textwidth}
        \tiny
        \textit{Note.} Numbers of coded segments in the KBD and control
        (CTRL) conditions are shown in brackets.
    \end{minipage}
\end{table}

The qualitative manipulation check illustrates behavioral differences
between conditions. Teachy Mini consistently used Socratic questions.
Hence, answers were provided by the student first, and Teachy Mini
provided additional explanations and more fine-grained follow-up
questions afterwards. On the other hand, the roles in the control
condition were switched: The student mainly asked questions, and the
robot provided explanations, which is generally considered a less
effective learning approach
\cite{elderRoleSocraticQuestioning1998,vandepolScaffoldingTeacherStudent2010}.
Relatedly, as the student in the KBD condition provided more answers,
Teachy Mini also gave more feedback by providing corrections or
affirmative statements (KBD: in 61.5\% of all messages; CTRL: in 9.8\%).
Among affirmative responses, Teachy Mini specified what exactly was
right in the students' explanations in almost all instances, whereas the
control robot used affirmative statements without explanation more often
(KBD: in 98.1\%; CTRL: in 9.8\%). Notably, in five of the 15 instances
where the students answered wrongly, Teachy Mini should have been more
assertive in correcting the students. However, Teachy Mini did not
provide any false information itself. On the other hand, the robot in
the control condition made ten statements that were factually wrong
(i.e., errors in 6.1\% of all messages), which highlights the danger of
LLM hallucination and false information in educational contexts
\cite{ciubotaruHallucinationProblemGenerative2025}. In addition, while
Teachy Mini explicitly linked 81.5\% of its claims to the lecture
slides, the robot in the control condition, which did not have this
contextual information, never provided sources. This limited the control
robot's transparency.

In the interactions we qualitatively investigated, Teachy Mini generally
adapted its behavior to the specified  KBD requirements. However, we were not able
to confirm whether the robot fulfilled all KBD requirements. To specify,
we did not identify reactions to user emotions, as students did not show
frustration in the interactions analyzed. Second, the robot did not
refer to the user's learning goal, possibly because it was not
particularly relevant in the current interaction paradigm. Further, the
Teachy Mini did relatively few adaptations to the students' learning
preferences. In two out of three interactions with Teachy Mini that were qualitatively analyzed, students reported preferences for humor in the opening questions. However, the robot only
used humorous explanations once. Although the KBD implementation guided
the robot's behavior in the desired direction, the robot's behavior was
still not fully consistent with its tasks, as exemplified by the fact
that Teachy Mini provided an affirmative statement without explanation
once, even though it was instructed to provide an explanation. As such,
while the robot may be aligned with its instructions in almost all
cases, the probabilistic nature of generative AI models still makes
truly robust behavior nearly impossible to achieve
\cite{benderDangersStochasticParrots2021,hundtLLMdrivenRobotsRisk2025}.
However, based on the clear behavioral distinction between conditions,
we deem the strength of the experimental manipulation strong enough for
preliminary investigations of student attitudes and learning gains.

\subsection{Analysis}\label{analysis}

Prior to hypothesis testing, we assessed the internal consistency and
item quality of the self-developed \emph{Perceived} \emph{Responsible
Behavior} scale. Factor loadings from an exploratory factor analysis,
corrected item-total correlations, Cronbach\textquotesingle s $\alpha$, and
McDonald\textquotesingle s $\omega$ were computed. The initial 10-item scale
demonstrated good internal consistency ($\alpha = .83$, $\omega = .92$).
However, inspection of corrected item-total correlations and factor
loadings revealed that item E3\_4 ($r = .232$, $\lambda =
.196$) fell below the recommended thresholds of $r = .3$ and
$\lambda = .3$ \cite{boatengBestPracticesDeveloping2018}. This item was
therefore excluded from the final scale. The resulting 9-item
\emph{Perceived} \emph{Responsible Behavior} scale demonstrated slightly
better internal consistency than the original scale ($\alpha = .86$,
$\omega = .93$).

We used descriptive statistics to investigate the validity of our
self-developed 12-item knowledge test. In the knowledge pretest,
participants had an average score of 8.42 (\emph{SD} = 2.19) from a
maximum of 12 points, indicating a ceiling effect. We then calculated
the percentage of correct answers (\emph{p}) for each pre- and posttest
item. On average, the items in the pretest were rather easy
(\emph{p\textsubscript{Total}} = 70.1\%, \emph{SD} = 23.8\%), with the
optimal difficulty for pretests being between 40\% and 60\%
\cite{boatengBestPracticesDeveloping2018,crockerIntroductionClassicalModern1986}.
Five items were very easy, with \emph{p} \textgreater{} 80\%:
\emph{p\textsubscript{D1}} = 87.5\%, \emph{p\textsubscript{D2}} =
95.8\%, \emph{p\textsubscript{D4}} = 83.3\%, \emph{p\textsubscript{D8}}
= 91.6\%, \emph{p\textsubscript{D10}} = 83.3\%. These five items were
excluded from analysis because they provided limited opportunity to
detect intervention-related learning gains. Hence, the final knowledge
test consisted of 7 items with an average pretest item difficulty of
57.1\% (\emph{SD} = 23.3\%).

Group differences for H1 (system acceptance), H2 (intrinsic motivation),
and H4 (alignment with responsible behavior) were examined using Welch
two-sample t-tests. This approach does not assume equal variances across groups. For H3a, objective learning effectiveness was analyzed using an
analysis of covariance (ANCOVA) with pretest scores as a covariate to
control for pre-existing knowledge differences between conditions. Regarding H3b, subjective learning effectiveness (perceived
competence) was analyzed using a Welch two-sample t-test.

An a priori power analysis with the G*Power software indicated that a
sample of 42 participants would be required to detect a large effect of
\emph{d} = .80 with $\alpha = .05$ and a statistical power of
$1-\beta = .80$ for between-group \emph{t}-test comparisons. Although
research suggests that Welch's \emph{t}-test is robust against Type I
error inflation in relatively small samples
\cite{delacreWhyPsychologistsShould2017}, the present sample (N = 24)
was underpowered and therefore had an increased risk of Type II errors.
Therefore, the results should be interpreted as preliminary. In
particular, non-significant effects should not be interpreted as
evidence for the absence of an effect.

\section{Results}\label{results}

The goal of the experiment was to investigate whether Teachy Mini would
lead to increased student acceptance (H1), motivation (H2), learning
effectiveness (H3), and perceived alignment with responsible robot
behavior (H4). First, we present these main hypothesis tests comparing
the KBD condition. Second, we report exploratory analyses that examine
potential covariates and relationships between the dependent variables.
These analyses provide a preliminary picture of whether KBD affected
students' perceptions, learning experience, and interaction outcomes.

\subsection{Main Hypotheses}\label{main-hypotheses}

A series of two-sample t-tests and an ANCOVA were conducted to
examine the four hypotheses. To test Hypothesis H1, a Welch two-sample t-test was
conducted comparing system acceptance between the KBD ($M = 5.04$,
$SD = 1.61$) and control ($M = 5.01$, $SD = 1.10$)
conditions. The test revealed no significant difference between
conditions, $t(19.43) = 0.057$, $p = .955$, $95\% CI =
\{-1.148, 1.212\}$. H1 was therefore not supported.

Hypothesis H2 was also tested using a Welch two-sample t-test comparing
intrinsic motivation between conditions. Although the KBD group
($M = 5.30$, $SD = 1.27$) reported descriptively higher
enjoyment than the control group ($M = 4.88$, $SD = 1.41$),
this difference did not reach statistical significance, $t(21.76)
= 0.762$, $p = .454$, $95\% CI = \{-0.718, 1.551\}$. Hence, H2 was
not supported.

Objective H3a was examined using ANCOVA with pretest score as a
covariate. The overall model was significant, $F(2, 21) = 4.506$,
$p = .024$, explaining 23.4\% of the variance in posttest
scores. Pretest score was a significant predictor of posttest
performance, \emph{B} = 0.394, \emph{t} = 2.952, \emph{p} = .008,
\emph{CI} = {[}0.116, 0.671{]}. The KBD group showed a mean learning
gain of 0.75 points (\emph{SD} = 1.54), compared to 0 points (\emph{SD}
= 1.54) in the control group. However, this effect was not significant,
\emph{B} = 0.447, \emph{t} = 0.967, \emph{p} = .344, $95\% CI =
\{-0.514, 1.407\}$. Additionally, we assessed H3b by comparing
participants' perceived competence between conditions. However, no
significant difference was found between the KBD group and the control
group, \emph{t}(19.80) = 0.31, \emph{p} = .762. By contrast, the control
group (\emph{M} = 4.71, \emph{SD} = 1.00) had descriptively higher
perceived competence than the KBD group (\emph{M} = 4.56, \emph{SD} =
1.41). As such, H3 was supported for neither objective nor subjective
learning effectiveness.

Regarding Hypothesis H4, a Welch two-sample t-test revealed a
significant difference between conditions. The KBD group (\emph{M} =
5.49, \emph{SD} = 1.23) reported significantly higher perceived
alignment with responsible robot behavior than the control group
(\emph{M} = 4.47, \emph{SD} = 0.95), \emph{t}(20.65) = 2.277, \emph{p} =
.034, \emph{CI} = {[}0.087, 1.950{]}, \emph{d} = 0.930. H4 was therefore
supported.

\subsection{Exploratory Analysis}\label{exploratory-analysis}

To investigate potential confounding variables and mediation paths, we conducted a series
of post-hoc exploratory analyses. First, as an alternative to the
\emph{t}-tests, we conducted linear regressions to explore whether
potential confounding variables affected our results. As such, for each
hypothesis (H1--H4) we included gender, age, level of education,
previous interactions with social robots, perceived baseline competence,
general learning motivation, group learning preference, and general
attitudes towards social robots as additional covariates. Only learning
preference had a significant impact on objective learning effectiveness,
such that participants who preferred learning in groups compared to
learning alone had significantly greater learning gains (\emph{B} =
0.667, \emph{t} = 2.251, \emph{p} = .042, \emph{CI} = {[}0.027,
1.308{]}). Further, after accounting for these covariates, the
coefficient for the KBD condition was positive and statistically
significant (\emph{B} = 1.714, \emph{t} = 2.355, \emph{p} \textless{}
.05, \emph{CI} = {[}0.142, 3.285{]}).

We then continued to investigate relationships between dependent
variables through Pearson correlations (Table \ref{tab5}). We identified
significant positive correlations between acceptance and motivation
(\emph{r} = .692, \emph{p} \textless{} .001), acceptance and subjective
learning effectiveness (\emph{r} = .729, \emph{p} \textless{} .001),
acceptance and perceived responsible behavior (\emph{r} = .512, \emph{p}
= .010), motivation and subjective learning effectiveness (\emph{r} =
.727, \emph{p} \textless{} .001), as well as motivation and perceived
responsible behavior (\emph{r} = .588, \emph{p} = .003).

\begin{table}[ht!]
    \centering
    \caption{Correlations among Study Variables}
    \label{tab5}
    \renewcommand{\arraystretch}{1.15}
    \setlength{\tabcolsep}{6pt}

    \begin{tabularx}{\textwidth}{
        >{\raggedright\arraybackslash}X
        *{5}{>{\centering\arraybackslash}p{0.075\textwidth}}
    }
        \toprule
        \textbf{Variable} &
        \textbf{1} &
        \textbf{2} &
        \textbf{3} &
        \textbf{4} &
        \textbf{5} \\
        \midrule

        1. Acceptance &
        --- & & & & \\

        2. Motivation &
        .692*** &
        --- & & & \\

        3. Objective Learning Effectiveness &
        $-.276$ &
        $-.149$ &
        --- & & \\

        4. Subjective Learning Effectiveness &
        .729*** &
        .727*** &
        $-.309$ &
        --- & \\

        5. Perceived Responsible Behavior &
        .512* &
        .588** &
        $-.153$ &
        .346 &
        --- \\

        \bottomrule
    \end{tabularx}
\end{table}

\begin{figure}[ht!]
\begin{center}
\includegraphics[width=\textwidth]{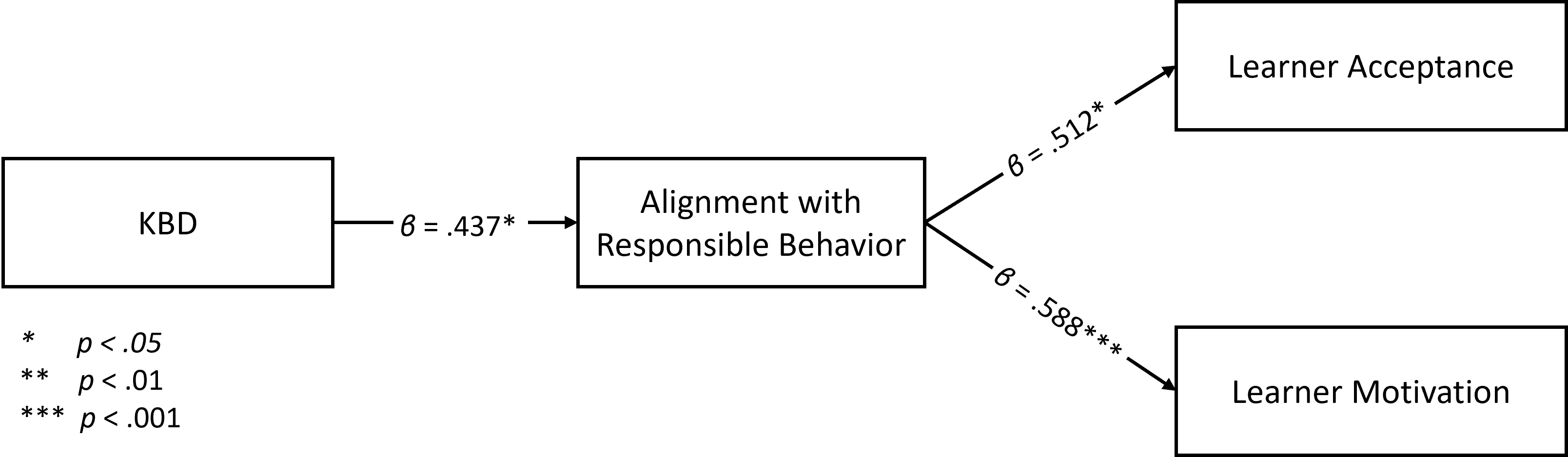}
\end{center}
\caption{Proposed Path Mode}
\label{Figure 4}
\end{figure}

Based on these intercorrelations between motivation, acceptance, and
alignment with responsible behavior, we further explored a potential
path model (Figure \ref{Figure 4}). In this model, perceived alignment with responsible behavior
was considered a predictor of both motivation and acceptance because a
robot that behaves transparently, appropriately, and responsively could
be experienced as a more trustworthy and pedagogically supportive
learning partner \cite{vonschallenKnowledgeBasedDesignRequirements2026}.
We kept the path from KBD to perceived alignment with responsible
behavior, as this path was supported by our main analysis. The proposed
model was tested as an observed-variable path model using structural
equation modelling, with all constructs represented by their respective
mean scores. The path analysis yielded fit indices conventionally
associated with acceptable or good fit ($\chi^2(2) = 2.197$, \emph{p} =
.333, \emph{CFI} = .991, \emph{TLI} = .972, \emph{RMSEA} = .059,
\emph{SRMR} = .074), but it should be interpreted cautiously as
exploratory due to the small sample size
\cite{maasSufficientSampleSizes2005}. The path from KBD to perceived
alignment with responsible behavior was significant ($\beta = .437$,
\emph{b} = 1.019, \emph{p} = .017, 95\% \emph{CI} {[}0.179, 1.858{]}).
Perceived alignment with responsible behavior, in turn, significantly
predicted motivation ($\beta = .588$, \emph{b} = 0.655, \emph{p}
\textless{} .001, 95\% \emph{CI} {[}0.374, 0.936{]}) and acceptance
($\beta = .512$, \emph{b} = 0.582, \emph{p} = .021, 95\% \emph{CI}
{[}0.068, 1.078{]}). There were also significant indirect effects of KBD
on learner acceptance ($\beta = .224$, \emph{b} = 0.593, \emph{p} =
.016, 95\% \emph{CI} {[}0.112, 1.073{]}) and motivation ($\beta =
.257$, \emph{b} = 0.667, \emph{p} = .019, 95\% \emph{CI} {[}0.109,
1.225{]}). The model explained 19.1\% of the variance in perceived
alignment with responsible behavior, 34.6\% of the variance in
motivation, and 26.3\% of the variance in acceptance.

\section{Discussion}\label{discussion}
The present reserach developed a GSR tutoring prototype that operationalized
selected KBD requirements for higher education and conducted a
preliminary evaluation study involving students and recent graduates.
Building on the KBD framework proposed by Vonschallen et al.
\cite{vonschallenKnowledgeBasedDesignRequirements2026}, we developed Teachy Mini, a robot prototype with integrated \emph{self-knowledge}, \emph{user-knowledge}, and
\emph{context-knowledge}. In a between-subjects study that featured a
robot-supported learning interaction, we compared Teachy Mini with a
reduced-knowledge control condition using the same robot platform and
language model. The main goal was not to establish definitive
educational effectiveness, but to examine whether KBD can be implemented
in a functioning robot tutor and to explore whether this configuration
can change students' perceptions and interaction experiences within a
single tutoring session.

The strongest finding concerned perceived alignment with responsible
robot behavior. Students in the KBD condition rated the robot as
significantly more aligned with responsible tutoring behavior than
students in the control condition. This finding provides preliminary
support for the core assumption of KBD: that responsible behavior in
GSRs can be shaped by systematically configuring what the robot knows
about itself, the learner, and the educational context
\cite{vonschallenKnowledgeBasedDesignRequirements2026}. This is
consistent with both educational AI and educational GSR research
emphasizing that content grounding, transparency, and personalization
are central for trustworthy learning support
\cite{kasneciChatGPTGoodOpportunities2023,tiszaGenerativeAIpoweredSocial2025}.

The results for system acceptance, intrinsic motivation, and learning
effectiveness remained inconclusive. Hypothesis H1 was not supported, as
system acceptance was nearly identical across conditions. Apart from
limitations regarding statistical power, one explanation is that during
one short interaction, differences in acceptance are harder to detect
and may be impacted by factors such as mere exposure or novelty effects
\cite{donnermannApplicationSocialRobots2025,reimannSocialRobotsWild2024}.
Another possibility is that KBD may have been indirectly associated with
acceptance through perceived alignment with responsible behavior. This
was suggested by the exploratory analysis of a potential path model and
is also consistent with prior research suggesting that social qualities
such as trustworthiness and pedagogical appropriateness can shape
acceptance of social robots
\cite{heerinkAssessingAcceptanceAssistive2010}.

Regarding hypothesis H2, intrinsic motivation was descriptively higher
in the KBD condition, but not by a significant amount. This directional
pattern is consistent with prior work linking personalization and
engagement to motivational outcomes in robot tutoring and AI-based
learning
\cite{paiEffectivenessSocialRobots2024,tasdelenGenerativeAIClassroom2025}.
Analogous to the findings on acceptance, our exploratory analysis suggests
that the degree of perceived alignment with responsible behavior may
have mediated the relationship between KBD and intrinsic motivation.
This pattern is theoretically plausible, as intrinsic motivation is
closely related to perceived competence and trust
\cite{ryanIntrinsicExtrinsicMotivations2000}. However, future research
is needed to investigate the potential of responsible, trustworthy GSRs
to increase learner motivation.

Hypotheses H3a and H3b were not supported in the main analyses for
either objective or subjective learning effectiveness. However,
objective learning gains were descriptively higher in the KBD condition,
and an exploratory covariate analysis showed a significant positive
effect of KBD on objective learning gains when accounting for
confounding variables such as preference for group learning. This
indicates that some students might benefit more from robot-supported
tutoring than others, which highlights the need for more customization
\cite{vonschallenKnowledgeBasedDesignRequirements2026}. The mixed
pattern regarding the effects of KBD on learning gains may also be
explained by the single-session nature of the experiment. In previous
research, adaptive robot-supported tutoring had positive effects on
learning gains in long-term interaction, but adaptive configurations did
not consistently outperform less adaptive versions within a single
interaction
\cite{donnermannApplicationSocialRobots2025,donnermannSocialRobotsApplied2022}.
Hence, more pronounced effects of KBD on learning effectiveness may only
emerge after repeated exposure to the robot. In real-world interactions
over longer periods of time, there may also be an indirect effect
involved, where stronger perceptions of responsible behaviors may
encourage learners to use the GSR more often -- as was indicated by our
exploratory mediation analysis. More frequent use of tutoring GSRs could
then lead users to learn more in general and, subsequently, induce
greater learning gains. This highlights the need for long-term field
research to increase our understanding of how robots impact learners in
the wild \cite{sabanovicRobotsWildObserving2006}.

Another interesting finding is that subjective and objective learning
effectiveness had a negative (although non-significant) relationship,
and students in the control group had descriptively higher average
perceived competence, but lower objective learning gains. While this
finding may be a statistical artifact from the low sample size, it may
also be explained by a key behavioral difference between the two
conditions: Teachy Mini focused on Socratic questions, while the robot
in the control group provided explanations only, which may reduce
critical thinking \cite{liCognitiveImpactChatGPT2026}. This echoes
broader concerns that students learning with generative AI technologies
may develop an inflated perception of their own understanding because
readily available explanations reduce the need for active knowledge
construction \cite{kasneciChatGPTGoodOpportunities2023}. Hence, GSRs
that are not designed to foster active learning, but just to deliver
content, may lead students to be overconfident in their own
understanding of the subject matter.

Lastly, the manipulation check illustrated that the behavior
of GSRs can be guided in a desired direction, but it cannot be
anticipated with complete certainty, echoing research that highlights
the stochastic nature of generative AI
\cite{benderDangersStochasticParrots2021,parkGenerativeAgentsInteractive2023,qiuBayesianTeachingEnables2026,yampolskiyAIUnexplainableUnpredictable2024}.
For example, in the KBD condition, the robot was instructed to affirm
user statements with an explanation about why the statement was correct.
However, despite doing so consistently in almost all interactions we
analyzed, it did not provide an explanation in one instance.
Furthermore, the robot in the KBD condition did not always correct the
students' wrong statements, despite being instructed to do so. This may
be related to known limitations of AI models regarding their sycophantic
behavior
\cite{chenWhenHelpfulnessBackfires2025,chengSycophanticAIDecreases2026,sunBeFriendlyNot2026}:
If students receive validation from GSRs for statements that are not
factually correct, their learning progress may be undermined.

\section{Strengths and Limitations}\label{strengths-and-limitations}

The current work offers several conceptual, technical, and
methodological strengths, while also being subject to limitations that
shape the interpretation and generalizability of its findings. A key strength of this research is that it provides the first integration
and preliminary experimental evaluation of KBD in a GSR. While previous
work identified KBD requirements conceptually and qualitatively
\cite{vonschallenKnowledgeBasedDesignRequirements2026,vonschallenKnowledgebasedDesignRequirements2026a},
the present study translated selected requirements into a functioning
tutoring robot prototype and examined their effects in an actual
learning interaction. This is important for responsible robotics because
it moves the discussion from abstract design principles toward
implementable system configurations that can be empirically evaluated.
However, the present work only investigated a selected subset of the
previously identified KBD requirements by Vonschallen et al.
\cite{vonschallenKnowledgeBasedDesignRequirements2026}. Although
requirements related to learning materials, personalization, educational
strategies, role, friendliness, assertiveness, and conscientiousness
were directly implemented, other requirements, such as long-term
learning progress, course schedules, grades, richer emotion recognition,
and awareness of the physical learning environment, were not implemented
or only partially addressed. In addition, a fully developed version of a robot with KBD should be more customizable by students
\cite{vonschallenKnowledgeBasedDesignRequirements2026} and moderated by
lecturers \cite{cordova-esparzaAIpoweredEducationalAgents2025}. The
limited scope of the prototype was appropriate for an initial controlled
prototype study, but it means that the evidence should also be
interpreted as a partial KBD implementation rather than an
implementation of the complete framework.

In the future, a tutoring robot based on KBD should also be evaluated in
longitudinal settings. This is particularly important as effects on
motivation, acceptance, and learning may require repeated exposure to the robot over
time. In addition, novelty effects may have influenced participants'
responses in the current evaluation
\cite{donnermannApplicationSocialRobots2025,reimannSocialRobotsWild2024},
particularly because almost all participants had no prior experience
interacting with social robots. Long-term field studies will be needed
to examine whether KBD effects persist after the novelty of the robot
has decreased and whether perceived responsible behavior translates into
sustained motivational and learning benefits. These long-term studies
will also be necessary to investigate how tutoring robots can be used as
social catalysts to motivate, rather than replace interactions with
teachers and other students
\cite{chenSocialRobotsConversational2025,orhaniRobotsAssistReplace2023,vonschallenKnowledgeBasedDesignRequirements2026}.

Another strength of the study is the controlled comparison between two
versions of the same robot platform. Both conditions used the same
Reachy Mini hardware, language model, learning material, session
duration, and general study procedure. This design helped isolate the
effect of the KBD integration more clearly than a comparison between
different platforms or between robot and non-robot conditions. The use
of a physical robot and authentic lecture materials further increased
the ecological relevance of the study, as participants interacted with
the system in a concrete learning task rather than in a short
demonstration or abstract usability scenario. The robot was used
alongside a set of available lecture slides. We did so because the
primary KBD use case identified for higher education was to help
students learn lecture contents
\cite{vonschallenKnowledgeBasedDesignRequirements2026}. Hence, having
lecture slides available increased the experimental realism of our
study. The slides were available in both the KBD and control conditions
-- which means that differences between conditions should still be
attributable to KBD. However, it is unclear whether the overall learning
progress stemmed predominantly fromthe interaction with the robot or
from self-engagement with the lecture slides. As a result, the
additional variance introduced by the lecture slides may have reduced
the strength of our experimental manipulation. Future studies should
investigate passive control conditions as well, such as learning without
a robot and with lecture slides only.

One of the major methodological limitations of the current study was the
limited sample size. Because the study was underpowered, the
non-significant findings for system acceptance, intrinsic motivation,
and learning effectiveness should not be interpreted as evidence that
KBD has no effect on these variables. Rather, they indicate that larger
studies are needed to estimate the effects of KBD more reliably.
Nonetheless, our study still found a significant effect for the key
assumption that KBD leads to more perceived alignment with responsible
behavior. Furthermore, when accounting for confounders such as learner
preferences, we observed a positive effect of KBD on learning gains. In
addition, the exploratory analysis provided important theoretical and
methodological insights for future confirmatory research, particularly
regarding potential mediation paths.

Another methodological limitation concerns the fact that our implementation relies heavily on
the OpenAI Realtime model. We chose this model for practical reasons
because it was, at the time, one of the few models capable of adaptive
open-ended communication in real time. Research indicates that different
LLMs with the same configuration may be somewhat comparable in terms of
their outputs
\cite{smithComprehensiveAnalysisLarge2025,vonschallenNeverSayNever2026}.
Still, it remains uncertain whether the observed effects of KBD truly
generalize to other AI models such as Gemini Live. With increasing model
capabilities, future research should also employ locally run models for
human-robot interaction research, which has advantages regarding both
privacy control and replicability
\cite{bakshOpensourceRoboticStudy2024,gunesReproducibilityHumanRobotInteraction2022}.

\section{Conclusion}\label{conclusion}

The paper provides an initial step toward empirically examining KBD for GSRs in higher education. Building on
previously identified KBD requirements, we demonstrated how combinations of
\emph{self-knowledge}, \emph{user-knowledge}, and
\emph{context-knowledge} can be operationalized in a functioning robot
tutoring prototype. The preliminary evaluation suggests that such
knowledge integration can make responsible tutoring behavior more
visible to learners, especially through personalization, slide
grounding, Socratic scaffolding, and learner-anchored feedback. While
the evidence for effects on motivation, acceptance, and learning remains
inconclusive, the study demonstrates that KBD can be treated as an
empirically testable design approach rather than only as a conceptual
framework. Future research can build on this by testing more complete
KBD implementations with larger samples and repeated interactions.

From an applied perspective, the findings are relevant for the
responsible development of educational robots and LLM-based tutoring
systems. As generative AI becomes increasingly integrated into learning
technologies, the central question is not only whether such systems can
produce fluent dialogue, but whether they can be configured to support
learners in accurate, transparent, motivating, and autonomy-preserving
ways
\cite{cordova-esparzaAIpoweredEducationalAgents2025,kasneciChatGPTGoodOpportunities2023,vonschallenKnowledgeBasedDesignRequirements2026}.
To achieve this, it is important to include students and lecturers in
the design process to identify relevant design requirements that align
with user needs \cite{vonschallenKnowledgeBasedDesignRequirements2026}.
Our Teachy Mini system was built on such design requirements and grounded
in explicit knowledge about its role, the learner, and the educational
context. Although further development and validation are needed before
deployment in real educational settings, KBD offers a practical
orientation for building GSR tutors that support students without
replacing human teachers, peer learning, or critical thinking.

Lastly, from a user-focused perspective, it is important to inform
students about how to responsibly interact with educational generative
AI technologies such as GSRs
\cite{hacklAILiteracyHeptagon2026,pangestuScopingLiteratureReview2026,tziridesCombiningHumanArtificial2024}.
This includes equipping students with the skills needed to critically
assess AI outputs, recognize their limitations, and use them in ways
that enhance rather than undermine independent learning and critical
thinking
\cite{cordova-esparzaAIpoweredEducationalAgents2025,elderRoleSocraticQuestioning1998,holmesEthicsAIEducation2022}.
One practical implication is that students may benefit from prompting AI
tools to use comprehension questions rather than relying exclusively on
direct explanations. In this sense, the future of educational AI is
unlikely to depend solely on building systems that know and explain
more, but on systems that support students in thinking critically.

\bibliographystyle{plain} 
\bibliography{references}

\end{document}